\documentclass{article}

\usepackage[final]{corl_2020} %

\usepackage{graphicx}
\usepackage{dsfont}
\usepackage{booktabs}
\usepackage{amsmath,amssymb}
\usepackage{tikz}

\newcommand{\V}[1]{{\mathbf{#1}}}
\newcommand{\M}[1]{{\ensuremath\mathbf{#1}}}
\newcommand{\attr}{\ensuremath \tau}
\newcommand{\attrs}{\ensuremath\mathcal{T}}

\newcommand{\dataset}{\textbf{SDVScenes}} %

\title{Universal Embeddings for Spatio-Temporal \\Tagging of Self-Driving Logs} 

\author{
  Sean Segal$^{12}$, Eric Kee$^{1}$, Wenjie Luo$^{12}$, Abbas Sadat$^{1}$, Ersin Yumer$^{1}$, Raquel Urtasun$^{12}$\\
  Uber Advanced Technologies Group$^{1}$, University of Toronto$^{2}$\\
  \texttt{\{ssegal,asadat,yumer,urtasun\}@uber.com} \\
}

\begin{document}
\maketitle

\begin{abstract}
In this paper, we tackle the problem of spatio-temporal tagging of self-driving scenes from raw sensor data. 
Our approach learns a universal embedding for all tags, 
enabling efficient tagging of many attributes and faster learning of new attributes with limited data.
Importantly, the embedding is spatio-temporally aware, allowing the model to naturally output spatio-temporal tag values. 
Values can then be pooled over arbitrary regions, in order to, for example, compute the pedestrian density \textit{in front of the SDV}, or determine if \textit{a car is blocking another car at a 4-way intersection}.
We demonstrate the effectiveness of our approach on a new large scale self-driving dataset, \dataset, containing $15$ attributes relating to vehicle and pedestrian density, the actions of each actor, the speed of each actor, interactions between actors, and the topology of the road map. 
\end{abstract}


\section{Introduction}
\label{sec:introduction}
\vspace{-0.1in}
In order to be deployed at scale, self-driving vehicles (SDVs) need to be extensively analyzed and tested in various challenging scenarios to ensure
proper handling of safety critical situations. 
Augmenting previous recordings of self-driving trips, or data logs, with rich metadata has many applications. For example, it enables efficient retrieval of scenarios for simulation, insightful failure analysis through visualization of tags most correlated with system failures, and rapid curation of better 
datasets for training the learned components of the system. 
Therefore, the ability to tag self-driving logs with useful 
metadata has become increasingly important for the development of SDVs.

Human experts are often used to label scenes with a variety of attributes to analyze failure modes, particularly in cases where the self-driving system disengaged or the safety driver took over. 
In the industry, this is known as {\it triage}. While this approach produces useful data, it does not scale. 
Alternatively, the outputs from the onboard perception system could be reused to heuristically reason about different scenarios based on the detections over time. 
Unfortunately, this approach requires extensive engineering, often returns noisy results, and cannot generalize to many scenarios which require reasoning beyond more than the detected actors.

Motivated by the shortcomings of these methods, in this paper we introduce a novel approach for spatio-temporal tagging of self-driving scenes, which requires only raw sensor data and HD maps as input and generalizes to tagging a diverse set of attributes.
Our  approach  learns  a  universal  embedding for all tags, enabling efficient tagging of many attributes in a given scene.
Given a particular tag attribute, we combine a learned attribute embedding with the universal data log embeddings to obtain a spatio-temporal tensor of attribute values.
This spatio-temporal tensor can be pooled over arbitrary regions, producing interpretable scene tags.
Our approach is trained end-to-end to minimize a multi-task tagging loss, with techniques to ensure learning is balanced across tags.

\begin{figure}[t]
    \vspace{-0.3cm}
    \centering
    \includegraphics[width=1\linewidth]{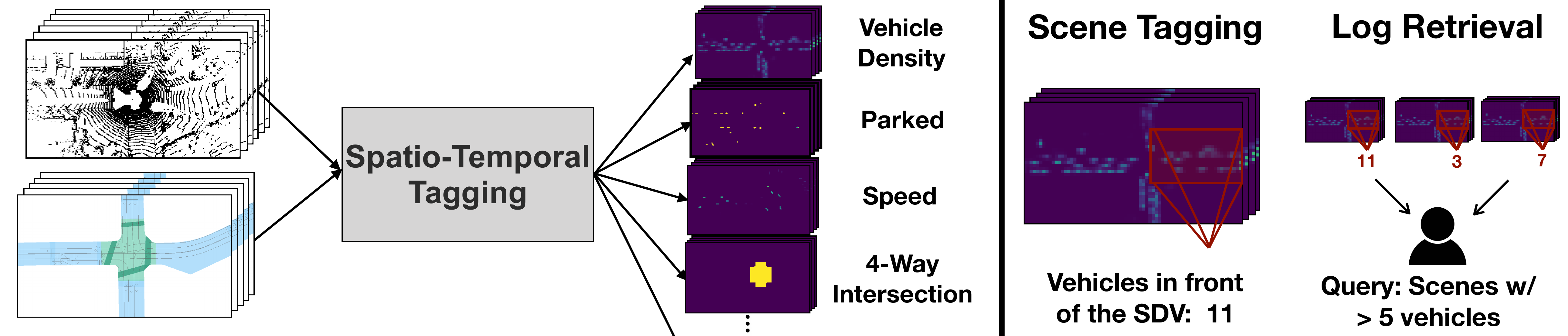}
    \caption{(Left) \textbf{Our task:} Tagging diverse attributes of a self-driving scene. (Right) \textbf{Applications:}  Spatio-temporal tags used downstream to generate interpretable scene tags and retrieve relevant logs.}
    \label{fig:our_task}
\end{figure}

Existing self-driving datasets either lack rich scene metadata \cite{geiger2013vision, sun2020scalability, chang2019argoverse, caesar2020nuscenes}, 
use only camera images \cite{malla2020titan}, 
or only provide annotations for the SDV itself rather than all actors in the scene \cite{xu2020explainable}. 
Therefore, we introduce \dataset, a novel large-scale dataset with over 40 hours of driving, containing LiDAR observations, HD maps and spatio-temporal annotations for 15 important scene attributes. Attributes are both discrete and continuous valued and cover actor density, vehicle actions, interactions, map topology information and vehicle speed for all actors. 

Using \dataset, we demonstrate that a single model can simultaneously
tag a diverse set of attributes relating to the self-driving scene. Additionally, we show that new attributes can be added at later training stages and achieve better performance than an independently trained model.
Finally, we analyze our system's performance with multiple ablation studies showcasing that our model leverages all inputs and is more efficient than separately trained models, while providing better performance. 
We plan to release a benchmark to encourage future work on this exciting new task.

\section{Related Work}
\label{sec:related}

\vspace{-0.1in}
\paragraph{Self-Driving Log Understanding:}
Few works address data log tagging or retrieval in the context of self-driving. 
\citep{lin2014visual} solves the problem of video retrieval 
using natural language queries by first parsing descriptions into semantic graphs and then matching them to visual concepts using a bipartite matching algorithm. 
This model, however, was trained on only 21 videos (8008 frames), whereas our dataset is several orders of magnitude larger. 
More recently, \citep{guangyu2019dbus} introduced a system to retrieve driving scenarios based on similarities in driver behavior using dash-cameras and IMU sensors. 
\citep{fu2019rekall} introduces a library for composing outputs of pretrained computer vision models for video analysis and demonstrates applications in the context of self-driving data log mining.

\vspace{-0.10in}
\paragraph{Video Understanding:} 
As the amount of video data continues to grow at a staggering rate, a rich literature has developed for tools to better understand and summarize this data. 
Many new network architectures (e.g., \citep{simonyan2014two, karpathy2014large, carreira2017quo, zhou2018temporal}) have been introduced to best capture both spatial and temporal relationships in videos for better classification and tagging. 
Many works additionally consider fusing video representations with language 
\citep{yang2003videoqa,zeng2017leveraging,tapaswi2016movieqa,zhu2017uncovering} 
to solve video question answering and captioning tasks. 
Work in video summarization introduced models which select either a subset of keyframes \citep{lee2012discovering,gong2014diverse} or keyshots \citep{lu2013story,ngo2003automatic,zhang2018retrospective,ji2019video}, which best preserve the information from the original video.

\vspace{-0.10in}
\paragraph{Action and Interaction Prediction:}
Self-driving scenes contain many actors, each performing one or more actions at any given time, such as turning, lane changing, or braking. As a result, 
many desired tags for self-driving scenes naturally relate to the current actions of agents in the scene.
Previous work has studied predicting the future actions, or intentions, of each agent in a scene. For example, \citep{streubel2014prediction, hu2018probabilistic} predict the intentions of vehicles at intersections. 
More recently, \citep{casas2018} introduced a model which jointly detects actors, classifies their intention and predicts their future trajectory from raw sensor data and HD maps. 
\citep{malla2020titan} predicts a set of hierarchically organized actions, used downstream as priors to better predict the future trajectories of actors and ego-motion. 
There has also been a growing interest in not only understanding each agent's current action, but also the interactions between agents in the scene \citep{kipf2018neural,Sadeghian2018SoPhieAA,Alahi2016SocialLH, li2020end, casas2019spatially, casas2020implicit}. 

\vspace{-0.10in}
\paragraph{Multi-task Learning:} 
Recent works in the self-driving domain have shown the ability to jointly learn multiple sub-tasks \citep{luo2018fast, kendall2018multi, zeng2019end, chowdhuri2019multinet, liang2019multi} for increased efficiency, and in certain cases better performance. 
More generally, there is increasing interest in techniques which balance conflicting multi-task objectives, including automatically reweighting losses \citep{kendall2018multi, chen2017gradnorm} and prioritizing specific tasks for efficient learning \cite{guo2018dynamic}. %

\section{Universal Embeddings for Spatio-Temporal Log Tagging}
\label{sec:model}
\vspace{-0.05in}

\begin{figure}[t]
    \vspace{-0.3cm}
    \centering
    \includegraphics[width=\linewidth]{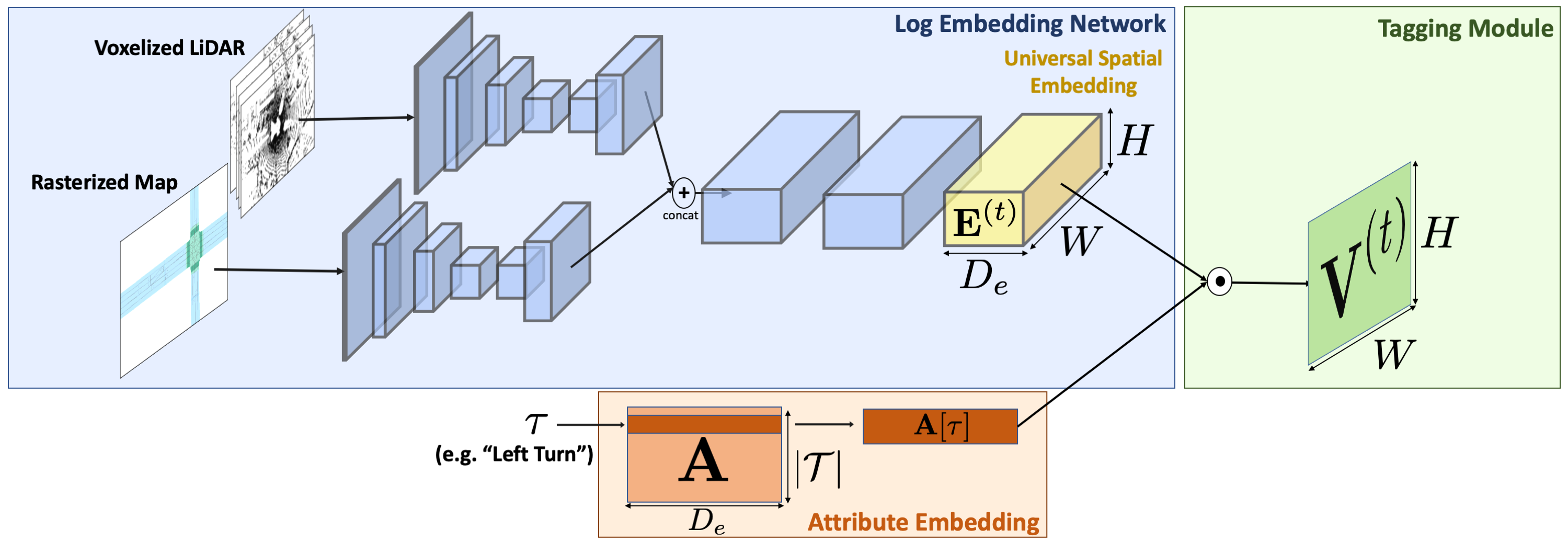}
    \caption{\textbf{Model design: } High level design of our model, 
    shown for one timestep, $t$.}
    \label{fig:model_main}
\end{figure}

Understanding raw data logs captured by self-driving vehicles is of crucial importance for applications such as simulation, triage analysis and dataset curation. 
In this paper, we tackle the problem of identifying precisely {\it when} and {\it where} complex events occur in raw data logs, 
represented by HD maps and LiDAR observations captured by a self-driving fleet. 
Towards this goal, we propose to learn a single universal embedding from raw logs capturing all information that is relevant to any of the supported tags. 
Importantly, this embedding maintains spatio-temporal dimensions: each element in the embedding represents information at a corresponding spatial location and timestep in the scene. 
Given a tag attribute, we lookup its learned embedding representation. 
Then, our tagging module takes as input the universal embedding and the attribute embedding, and extracts the relevant information to produce spatio-temporal tag values. 
Importantly, our model is trained end-to-end to both learn the embeddings and compute all tags. 
An overview of our approach is outlined in Fig.~\ref{fig:model_main}. 

We approach the problem of spatio-temporal tagging by learning a function, $f$, which takes a data log
$\M L$
and a tag attribute $\attr$ from a predefined set $\attrs$ and returns spatio-temporal tag values, 
\begin{equation}
\M V = f(\M L, \tau)~~.
\end{equation}
 $\M V$ maintains spatio-temporal dimensions, $T \times H \times W$. Each element $\M V^{(t)}_{h, w}$ is a value, either discrete or continuous, corresponding to a specific timestep, $t$ and spatial location in the observed scene, represented by the coordinates $(h, w)$ in Bird's Eye View. 
 Next, we describe our approach to representing $f$ for all tags $\attr \in \attrs$ with a single neural network. We also demonstrate extensions to compute a tag's value for an arbitrary region and compose  attributes for more complex tags.  

\vspace{-0.05in}
\subsection{Universal Embeddings}
\vspace{-0.05in}
\paragraph{Log Embedding Network:}
Our fully convolutional embedding network, $f_e^{\theta}$ takes a raw log   $\M L$   as input and outputs a spatio-temporal universal embedding:
\begin{equation}
	\M E = f_e^{\theta}\big(\M L\big)~~,
\end{equation}
where $\M L$ represents the recorded LiDAR and HD maps for the entire log. Note that $f_e^\theta$ does not depend on any particular tag attribute. This allows our model to efficiently share the computation of important intermediate features that may be relevant to multiple tags. Additionally, this enables our embedding to be precomputed and stored for fast on demand tagging.

Because data logs can have arbitrary length, $T$, our embeddings are fully convolutional  across the time dimension with a receptive field of $N$ timesteps. 
Having a sufficiently large temporal receptive field is important, as many attributes might require looking at several frames to be tagged accurately (e.g., braking, turning, vehicle speed). 
The $N$ \mbox{LiDAR} sweeps are corrected for ego-motion to bring the point clouds into the same coordinate system, centered at the SDV. 
We follow \cite{casas2018} and rasterize the space into a 3D occupancy grid, where each voxel indicates whether it contains a LiDAR point. 

Many tag attributes require reasoning about each actor's position with respect to the road map. For example, tagging lane changes requires understanding the position of each actor with respect to the  lanes. 
Following \cite{casas2018}, we rasterize the map into $M$ channels, each representing a different element, e.g., road, intersections, lanes, lane boundaries, traffic lights. The full input representation for a given frame is therefore a tensor of size $H_L \times W_L \times (ZN + M)$,  where $Z$ and $H_L$, $W_L$ are the height and x-y dimensions respectively. 
The embedding for each frame $\M E^{(t)}$, computed by $f_e^{\theta}$, has size $H \times W \times D_e$, where $D_e$ is the embedding dimension and the spatial dimensions $H, W$ are obtained by downsampling $H_L$ and $W_L$ by a factor of $r$.

The architecture of our network is inspired by recent works in object detection \citep{yang2018pixor}. 
First, we process the voxelized LiDAR and the rasterized map with independent backbones. 
Then, their features at multiple resolutions are upsampled and concatenated together, which is given to a convolutional header to obtain our universal embedding. Please refer to the supplementary material for details. 

\begin{figure*}[t]
    \centering
    \includegraphics[width=1.0\linewidth]{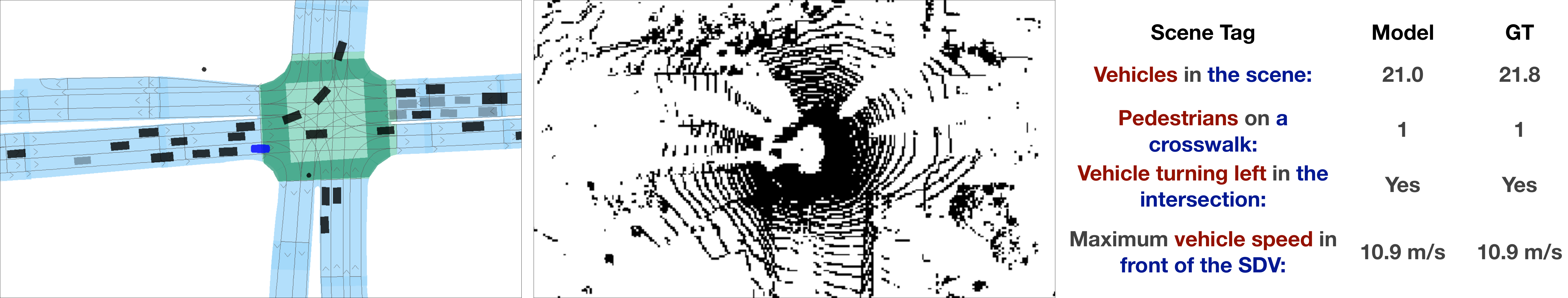}\
    \includegraphics[width=1.0\linewidth]{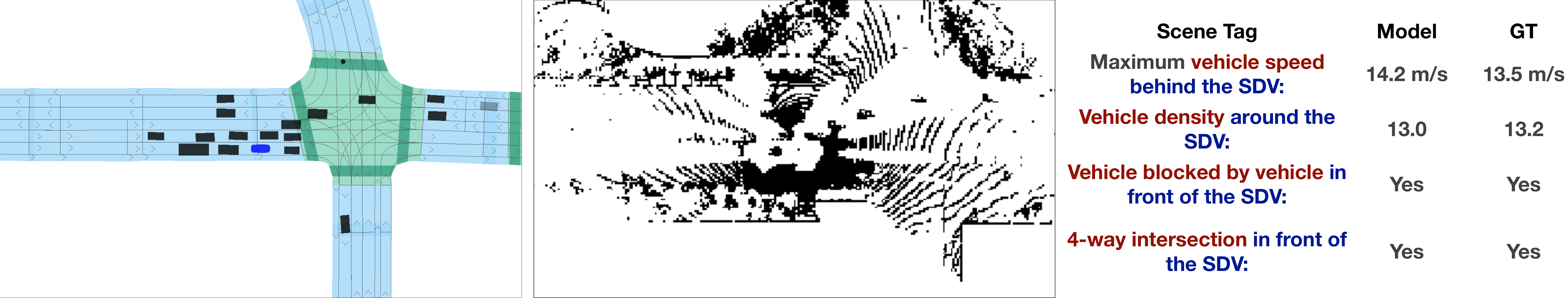}\ 
    \caption{\textbf{Qualitative Scene Tags:} (Left) Visualization of the scene (\textbf{not} given to the model). (Center) Sensor observations for a central timestep (Right) Scene tags generated by our model.}
    \vspace{-0.3cm}
    \label{fig:qualitative_search_results}
    \vspace{-0.1cm}
\end{figure*}

\vspace{-0.10in}
\paragraph{Attribute Embedding:}
Tag attributes  might be related to one another. 
For example, the action ``vehicle braking'' and interaction ``vehicle is braking due to another vehicle'' are clearly related. 
Other attributes, like turns and intersection types, may also be related in more complicated ways.
Inspired by the success of word embeddings in NLP \citep{mikolov2013distributed}, we learn an embedding representation for each tag attribute. 
Specifically, we introduce a learnable embedding matrix $\M A$ with dimension $|\attrs| \times D_e$, where $|\attrs|$ is the number of attributes. 
Given a tag attribute, $\attr$, its embedding is obtained by indexing the attribute's corresponding row, $A[\attr]$.

\vspace{-0.10in}
\paragraph{Tagging Module:} 
Given the log embedding representation $\M E$ and the attribute representation $\V a = A[\attr]$ as input, our tagging module returns a spatio-temporal tensor of tagged values with dimension $T \times H \times W$  
using a pointwise dot product along the embedding dimension,
\begin{equation}
\label{model:tagging} 
	\M V = \M E \odot \V a ~~.
\end{equation}
This approach is parameter free, forcing our learned embeddings to capture all relevant information.

\vspace{-0.10in}
\paragraph{Scene Tags:}
Our system can also compute a tag's value over an arbitrary region, $R$, via pooling,
\begin{equation}\label{model:pooling}    
    v = f_{p}(\M V^{(t)}, R)~~.
\end{equation}
These regions can be defined with respect to the SDV, for example encoding the region in front of, or behind the SDV, or according to regions from an HD map, for example, encoding regions with a crosswalk. For example, if $\M V^{(t)}$ is the pedestrian density at each spatial location, we can obtain an estimate of the region's density with $f_p = \texttt{sum}$, summing the values in $\M V^{(t)}$ over $R$.
We also use $f_p = \texttt{max}$ to pool logits, outputting the probability that an attribute is present {\it somewhere} in $R$.

\vspace{-0.10in}
\paragraph{Compositional Tags:} 
Many interesting scenarios in self-driving scenes can be best expressed in terms of simple scene attributes. For example, tagging scenes with vehicle's stopped at a 4-way intersection could be solved by recognizing stopped vehicles and 4-way intersections independently, and then reasoning about the existence of the two outputs to obtain a final tag. Motivated by this compositionality, we extend our model to support compositional tags, which can be expressed as functions of the outputs of simpler tags. More formally, we define a compositional tag $\attr_{c}$ as,
\begin{equation}\label{model:compositional_tag}
    f(\M L, \attr_{c}) = g\left(\{\M V_{\attr}: \attr \in \attrs_{c}\}\right)~~.
\end{equation}

Compositional attributes are fully defined by the compositional function $g$ and the subset of tag attributes $\attrs_{c} \subseteq \attrs$. In this work, we use simple compositional functions of the form,
\begin{equation}\label{model:compositional_function}
g_{\text{AND}}\left(\cdot\right) = \prod_{\attr \in \attrs_{c}} h_{\attr}(\M V_{\attr})~~,
\end{equation}
where $h_{\attr}(\cdot)$ is either the identity or an indicator function $\mathds{1}\{\M V_{\attr} \geq \kappa\}$ applied elementwise. As an example, imagine we have  two attributes ``left turn'' and ``4-way intersection'', which have taggers which output the probability of each event occurring at locations across the scene. 
Then, the output of the multiplicative compositional function $g_{\text{AND}}$ corresponds to the probability 
that there is a vehicle turning left at a 4-way intersection under the assumption that the events are independent. $g_{\text{OR}}$ can be derived similarly, using a sum instead of product and subtracting by $g_{\text{AND}}$ as the events may not be mutually exclusive. 
We leave the use of more advanced, potentially learned, compositional functions as future work.   

\begin{figure}[t]
        \centering
        \includegraphics[width=1.0\linewidth]{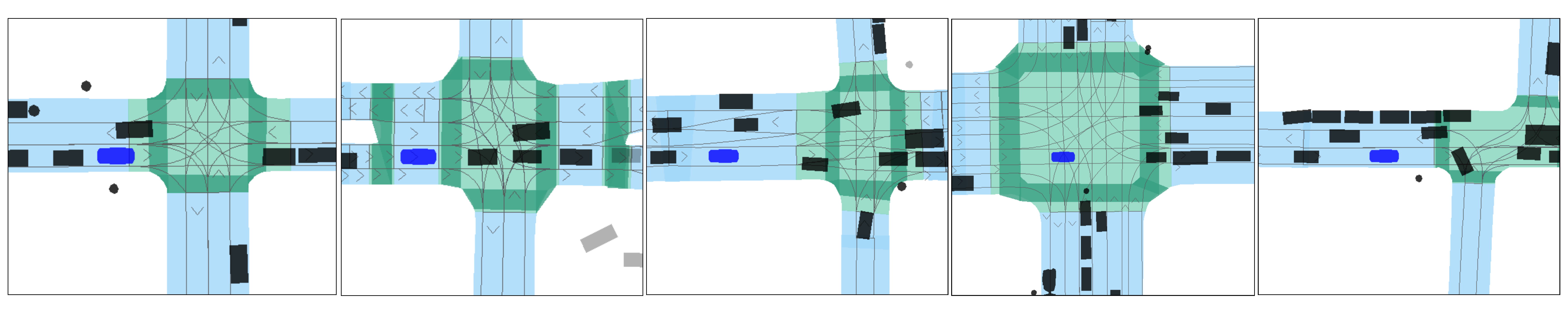}\
        \caption{
        \textbf{Log Retrieval:} Visualizations of scenes retrieved by our model using the compositional tag for vehicles blocked by vehicle at a 4-way intersection in front of the SDV. Importantly, our model is given only raw sensor data and HD maps, not the actor labels used for visualization.
        }
        \label{fig:retrieval}
\end{figure}

\vspace{-0.05in}
\subsection{Learning} 
\label{model:learning}
\vspace{-0.05in}
We use supervised learning  to jointly learn both the embedding network and the attribute embeddings end-to-end. 
Let $\Theta = \{\theta, \M A\}$ be the collection of model parameters. Given a database of training logs 
$\mathcal{L} = \{\M L\}$ 
with ground truth tag values for each attribute $\tilde{\M V}_{\attr}$, we train our model to minimize a multi-task tagging loss 
which minimizes the loss across all attributes, timesteps, and data logs in the training set, 
\begin{equation}
\label{model:loss}
\min_{\Theta} \sum_{\M L} \sum_{t}^{T} \left(\sum_{\attr \in \mathcal{D}_{\attr}} \ell_{d} (f_{\Theta}(\M L^{(t)}, \attr), \tilde{\M V}_{\attr}^{(t)}) + \sum_{\attr \in \mathcal{C}_{\attr}} \ell_{c} (f_{\Theta}(\M L^{(t)}, \attr), \tilde{\M V}_{\attr}^{(t)} ) \right)~~,
\end{equation}
where $\mathcal{D}_{\attr}$ and $\mathcal{C}_{\attr}$ are the discrete and continuous tag attributes, respectively. The loss for discrete attributes, $\ell_d$ is the standard cross entropy loss, applied per-pixel. For continuous-valued attributes, the loss $\ell_{c}$ is a standard regression loss (e.g., smooth L1) applied both independently per-pixel and, for density tags, after pooling via summation over the entire scene. Applying the loss after pooling ensures that errors will not be spatially correlated.
Naively optimizing Equation~\ref{model:loss} with these losses will be suboptimal given both large imbalances in the tagged value distributions and the multi-task nature of the objective. 
Therefore, we leverage two techniques to ensure stable and efficient learning.

\paragraph{Hard Negative Mining:}
In the context of self-driving, the most interesting scene attributes often occur infrequently. For example, interesting maneuvers, like lane changes and actor-to-actor interactions, will predominantly take on negative values, resulting in highly imbalanced supervision.
We solve this using hard negative mining and only apply the loss to
negative pixels in $\tilde{\M V}^{(t)}$ with the highest predicted confidences in ${\M V}^{(t)}$.
In practice, we sort the predictions of each pixel where the ground truth is negative by their predicted confidences
and only apply the cross entropy loss to the hardest examples, 
ensuring there are at most $K = 3$ negative pixels for each positive pixel in a frame.

\paragraph{Task Balancing:} Hard negative mining ensures that for each attribute, $\attr \in \attrs$, the distribution of the tagged values is balanced. At the same time, it introduces imbalance between the amount of supervision each task receives. As an example, consider two attributes which we use in our system: parked and left lane changes. Almost all frames will have at least one parked vehicle producing useful signal for the network, while the many frames without left lane changes will produce no signal.
To counter this imbalance, we preprocess our dataset and for each attribute, $\attr \in \attrs$, we compute the subset of frames that have at least one location with a positive label, $\mathcal{L}({\attr})$. Then, to construct minibatches for training, we first sample a tag attribute uniformly and then sample a log frame $\M L^{(t)}$ that is ``interesting'' for the given task uniformly from $\mathcal{L}(\attr)$. 

\setlength{\tabcolsep}{0.2em}
\renewcommand{\arraystretch}{1.08}
\begin{table}[t]
\vspace{-0.3cm}
\centering
\footnotesize
\begin{tabular}{@{\extracolsep{0.2em}}cccccccccccccccc@{\hspace{0.0em}}}

\\ \toprule
	
\multicolumn{1}{c}{} &
\multicolumn{2}{c}{Density {\scriptsize (L1) $\downarrow$} }  &
\multicolumn{10}{c}{Actions {\scriptsize (F1) $\uparrow$} }  &
\multicolumn{2}{c}{Map {\scriptsize (F1) $\uparrow$ }} &
\multicolumn{1}{c}{Speed {\scriptsize (L1)  $\downarrow$}} 
\\
	\cmidrule{2-3}
	\cmidrule{4-13}
	\cmidrule{14-15}
	\cmidrule{16-16}
	
	Region  &
	\text{\scriptsize Veh} &
	\text{\scriptsize Ped} &
	\text{\scriptsize P} &
	\text{\scriptsize S} &
	\text{\scriptsize B} &
	\text{\scriptsize KL} &
	\text{\scriptsize RT} &
	\text{\scriptsize LT} &
	\text{\scriptsize RC} &
	\text{\scriptsize LC} &
	\text{\scriptsize BB} &
	\text{\scriptsize BF} &
	\text{\scriptsize 3-way} &
	\text{\scriptsize 4-way} &
	\text{\scriptsize Speed}  
\\\midrule
Full &
$0.93$ &
$1.40$ &
$93$ &
$95$ &
$83$ &
$98$ &
$76$ &
$76$ &
$44$ &
$53$ &
$66$ &
$67$ &
$99$ &
$99$ &
$0.62$  
\\ 
Around &
$0.46$ &
$0.90$ &
$94$ &
$94$ &
$75$ &
$96$ &
$76$ &
$77$ &
$45$ &
$55$ &
$68$ &
$67$ &
$100$ &
$99$ &
$0.48$  
\\ 
Front &
$0.31$ &
$0.61$ &
$93$ &
$92$ &
$67$ &
$92$ &
$74$ &
$78$ &
$39$ &
$52$ &
$63$ &
$59$ &
$99$ &
$99$ &
$0.49$  
\\ 
Behind &
$0.29$ &
$0.45$ &
$94$ &
$89$ &
$71$ &
$93$ &
$76$ &
$72$ &
$46$ &
$56$ &
$69$ &
$65$ &
$100$ &
$99$ &
$0.46$  
\\ 
Intersection &
$0.31$ &
$0.27$ &
- &
$83$ &
$58$ &
$91$ &
$78$ &
$77$ &
$37$ &
$44$ &
$59$ &
$48$ &
$99$ &
$99$ &
$0.59$  
\\ 
Crosswalk &
- &
$0.21$ &
- &
- &
- &
- &
- &
- &
- &
- &
- &
- &
- &
- &
-  
\\

\bottomrule
	
\end{tabular}

\vspace{0.2cm}

\caption{\textbf{Scene Tagging:} Metrics computed over different regions of interest. Values are omitted if tag is not applicable (e.g., parked at an intersection)}
\label{table:spatial_metrics}
\vspace{-0.2cm}
\end{table} 
\begin{figure}[t]
  \vspace{-0.3cm}
  \centering
  \begin{minipage}{.45\textwidth}
\setlength{\tabcolsep}{0.04em}
\renewcommand{\arraystretch}{1.04}
\vspace{-0.3cm}
\centering
\footnotesize
\begin{tabular}{@{\extracolsep{0.6em}}cccccccc}

\\ \toprule
	
\multicolumn{1}{c}{} &
\multicolumn{2}{c}{Joint Embedding}  &
\multicolumn{2}{c}{Separate Embedding}  
\\
	\cmidrule{2-3}
	\cmidrule{4-5}
	
	$D_e$  &
	\scriptsize{Disc. (F1) $\uparrow$} &
	\scriptsize{Cont. (L1) $\downarrow$} &
	\scriptsize{Disc. (F1) $\uparrow$} &
	\scriptsize{Cont. (L1) $\downarrow$} &
\\\midrule
	4 &
	$50$ &
	$1.62$ &
	$10$ &
  $494^{*}$
\\ 
	16 &
	71 &
	$0.98$ &
	$50$ &
	$1.20$ 
\\ 
64 &
\textbf{72} &
\textbf{0.93} &
\textbf{71} &
\textbf{0.92} 
\\ 

\bottomrule
	
\end{tabular}
\vspace{0.2cm}

  \end{minipage}%
  \hspace{0.1cm}
  \begin{minipage}{.5\textwidth}
    \includegraphics[width=\linewidth]{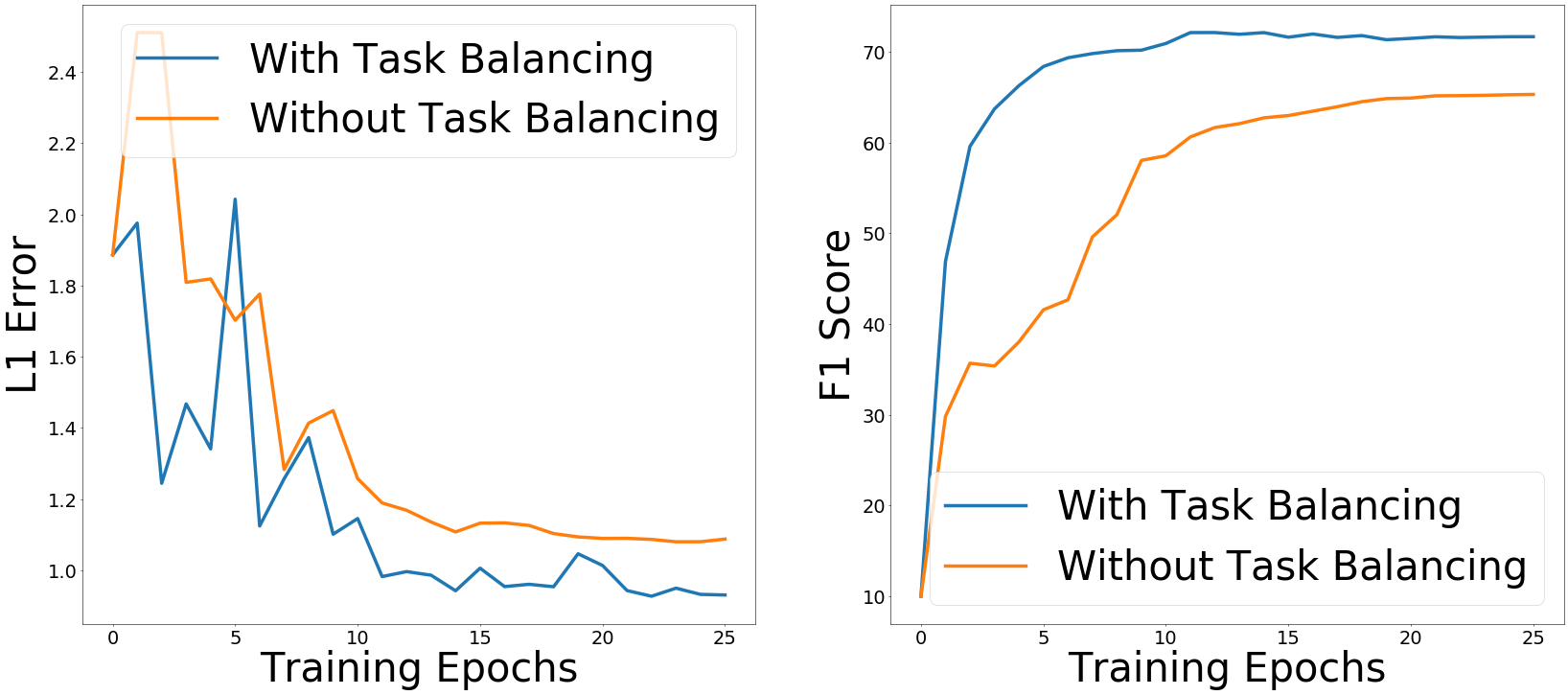}
  \end{minipage}
  \caption{(Left) \textbf{Embedding Ablations:} Geometric mean of metrics for  different embedding dimensions and configurations. *Model unable to learn given memory limitations. (Right) \textbf{Task Balancing:} Effect of our task balancing sampling scheme on learning.}
  \vspace{-0.3cm}
  \label{fig:dimension_ablation_and_task_balancing}
\end{figure}

\begin{figure}[t]
  \centering
  \begin{minipage}{.55\textwidth}
    \includegraphics[width=\linewidth]{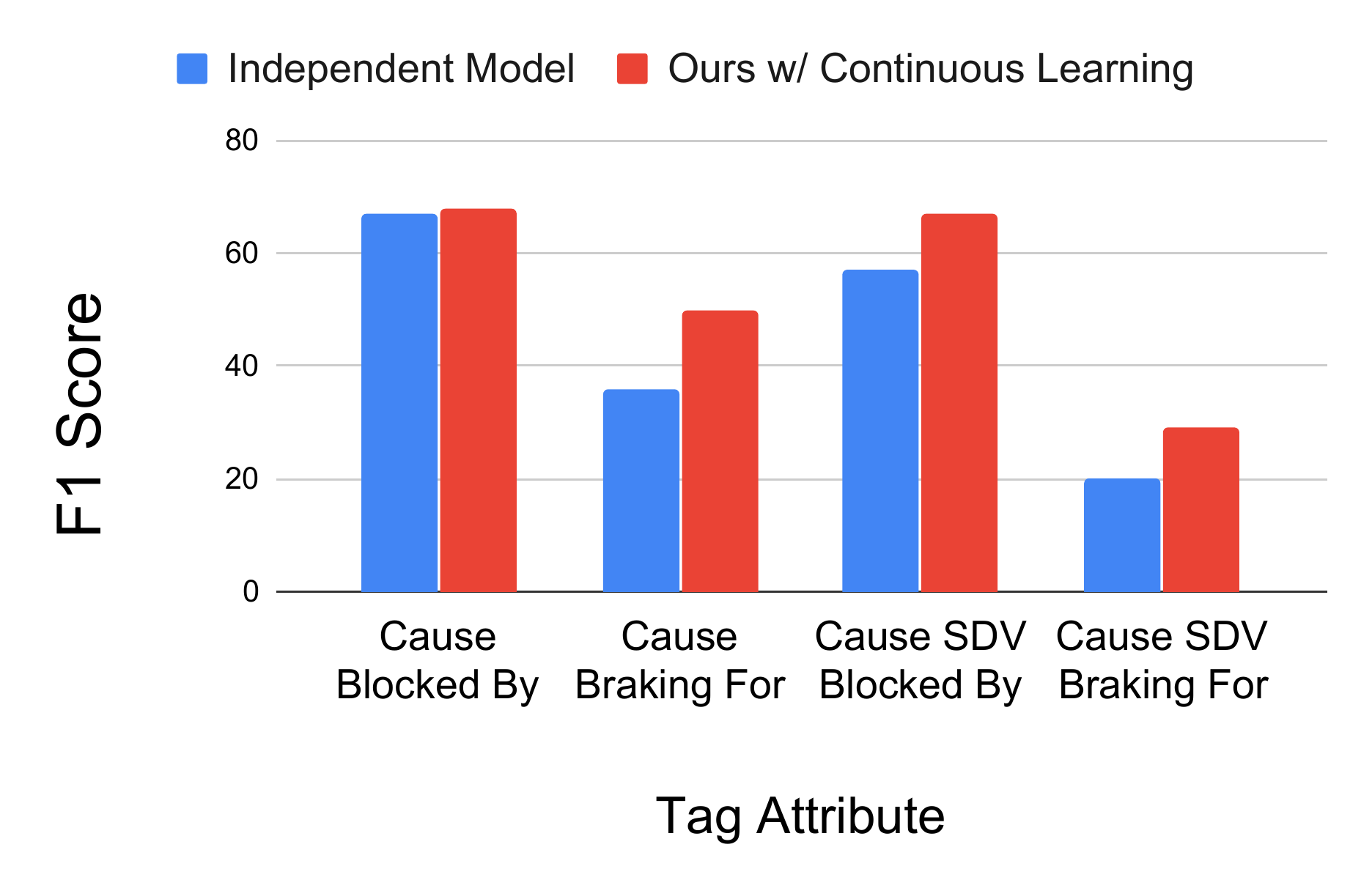}
  \end{minipage}%
  \begin{minipage}{.45\textwidth}
    \includegraphics[width=\linewidth]{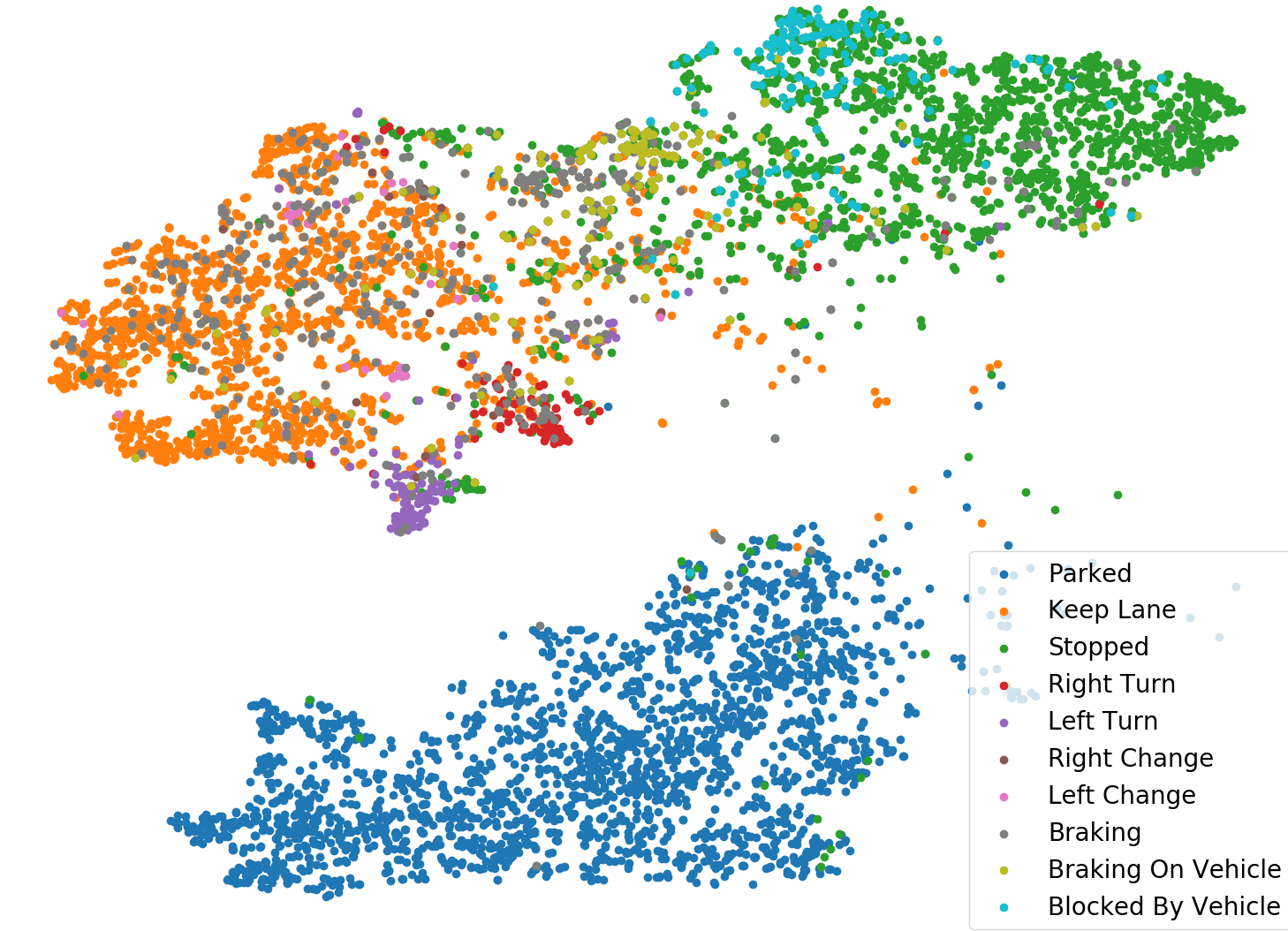}
  \end{minipage}
  \caption{(Left) \textbf{Continuous Learning:} New attributes on the x-axis. (blue)  independently trained baseline model, (red) continuously learned approach. (Right) \textbf{Embedding Visualization:} t-SNE~\cite{maaten2008visualizing} of the universal embedding space. Points correspond to embedding vectors at spatial locations in $\M E^{(t)}$ with vehicles. Colors indicate the primary action or interaction of the vehicle.}
  \vspace{-0.3cm}
  \label{fig:continuous_learning_tsne}
\end{figure}

\section{Dataset and Experiments}
\label{sec:experiments}
\vspace{-0.1in}

We evaluate our approach on our new large scale dataset, \dataset ~(see Section \ref{sec:dataset}), showcasing our model's ability to simultaneously tag a diverse and complex set of scene attributes. 
We also evaluate our model in the continuous learning setting, where new tag attributes are added at later stages of the training process. 
In particular, we show stronger performance when new attributes are learned from a pretrained universal embedding, demonstrating that information captured by our embedding generalizes to new attributes. 
Finally, we perform a set of ablation studies to understand how different inputs, embedding configurations \& sizes, and training schemes affect performance. 

\subsection{\dataset~Dataset}
\label{sec:dataset}
\vspace{-0.1in}
We introduce a novel large-scale dataset which contains sensor observations, HD maps, and annotations of $15$ scene attributes relating to actor density, actions, interactions, map topology and vehicle speed, 
which will be released in a future benchmark.

For the data logs, we collected roughly $40$ hours of driving over multiple cities across North America. We split the data into $4857$ data logs for training, $477$ for validation and $960$ for testing. 
Each log is roughly $25$ seconds, and our dataset provides observations and supervision at $10$ Hz,
resulting in roughly $1$M training frames, $100$K validation frames, and $200$K testing frames. 

\paragraph{Continuous Scene Attributes:} Our dataset contains annotations for 3 continuous-valued scene attributes, divided into two categories: actor speed and density.
We provide annotations for the speed of each moving vehicle, as it is an important aspect of the scene.
Our dataset also provides bounding box annotations for both vehicles and pedestrians, allowing us to compute the density for any region in the scene. 
We represent this actor density as a continuous attribute to  handle cases where vehicles on the boundary are only partially visible (e.g., 0.5 vehicle density). 

\paragraph{Discrete Scene Attributes:} Our dataset contains annotations for 12 discrete scene attributes. For each vehicle, 
we annotate $8$ common actions: Parked, Stopped, Braking, Keeping Lane, Right (Left) Turn, Right (Left) Change. 
Additionally, our dataset contains annotations for two vehicle-to-vehicle interactions:  whether a vehicle is blocked-by, or braking-for another vehicle. Precise definitions of these can be found in the supplementary material. 
All actions and interactions take binary values and are labeled independently. 
We also have annotations for the map's topology: intersections are either 3-way intersections, 4-way intersections or neither (intersections with more than 4 arms).

\paragraph{Implementation Details:} 
For all experiments, we use $H_L = 160$, $W_L = 280$ (at a $0.5$ meter per pixel resolution), $Z=3$ (at a 1 meter per pixel resolution), a spatial downsampling rate of $r = 2$ and $M = 15$ map channels. We use $N = 10$ frames, sampled at $5$ Hz. Unless otherwise noted, the embedding dimension is $D_e = 64$. 
For each epoch, we sample $25000$ examples per attribute without replacement and minimize the loss in Eq. (\ref{model:loss}). To train the parameters, we use the Adam optimizer~\cite{kingma2014adam} with an initial learning rate $\alpha = 0.0001$ that is decayed by $0.1$ every $10$ epochs. 
We use a batch size of $10$ examples per GPU and employ data-parallelism to train with 32 GPUs. 
We notice that density tags require more examples than others in order to reach convergence. Hence, we train our final model in two stages. 
First, we train on density tags for $30$ epochs. Then, we add all remaining tags and train for $25$ more epochs. 

\subsection{Tagging Performance}
\paragraph{Scene Tagging:} 
In Table~\ref{table:spatial_metrics}, we evaluate the performance of our model in the scene tagging setting for a variety of regions, $R$.
For regions, we use the entire scene, the region in front of, behind and around the SDV (each defined precisely in the supplementary).
Additionally, we evaluate on 2 map-based regions: intersections and crosswalks. 
Given that actions, interactions and speed tags are only trained on locations with vehicles, tagging one of these attributes $\attr$, over a larger region requires reasoning about both the presence of a vehicle and the value of $\attr$ at each location. 
Therefore, we use a compositional tag that depends on both vehicle density and the tag attribute $\attr$, where $g$ is a compositional function which thresholds the density to obtain locations with vehicles which is multiplied by $f(\M L, \attr)$ to obtain the final tensor of tagged values. 
We use L1 error and F1 score to measure the performance of continuous and discrete attributes, respectively.
Generally, tagging vehicles in intersections appears most challenging, 
likely because vehicle behavior is more complicated in these regions compared to regular roads. Fig.~\ref{fig:qualitative_search_results} shows sample scene tags output by our model. 

\paragraph{Retrieval Setting:} 
We test our model qualitatively in the retrieval setting by searching 
for ``gridlocked'' scenarios at intersections, known to be challenging for SDVs. 
For all logs in the evaluation set, we tag whether there is a vehicle blocked by vehicle at a 4-way intersection in front of the SDV. 
In Fig.~\ref{fig:retrieval}, we show visualizations of the scenes with actor labels shown (not given to the model, but useful for qualitative evaluation) that our model tagged as positives.
The results show a diverse set of scenes where the SDV is at a crowded intersection, which could be used in practice to test whether the SDV would plan a safe maneuver without blocking the intersection. 

\setlength{\tabcolsep}{0.1em}
\renewcommand{\arraystretch}{1.08}
\begin{table}[t]
\vspace{-0.3cm}
\centering
\footnotesize
\begin{tabular}{@{\extracolsep{0.2em}}ccccccccccccccccc@{\hspace{0.0em}}}
\tiny

\\ \toprule
	
	\multicolumn{2}{c}{Inputs} &
	\multicolumn{2}{c}{Density {\scriptsize (L1) $\downarrow$} }  &
	\multicolumn{10}{c}{Actions {\scriptsize (F1) $\uparrow$} }  &
	\multicolumn{2}{c}{Map {\scriptsize (F1) $\uparrow$ }} &
	\multicolumn{1}{c}{Speed {\scriptsize (L1)  $\downarrow$}} 
\\
	\cmidrule{3-4}
	\cmidrule{5-14}
	\cmidrule{15-16}
	\cmidrule{17-17}

	\text{\scriptsize Map}  &
	\text{\scriptsize LiDAR}  &
	\text{\scriptsize Veh} &
	\text{\scriptsize Ped} &
	\text{\scriptsize P} &
	\text{\scriptsize S} &
	\text{\scriptsize B} &
	\text{\scriptsize KL} &
	\text{\scriptsize RT} &
	\text{\scriptsize LT} &
	\text{\scriptsize RC} &
	\text{\scriptsize LC} &
	\text{\scriptsize BB} &
	\text{\scriptsize BF} &
	\text{\scriptsize 3-way} &
	\text{\scriptsize 4-way} &
	\text{\scriptsize Speed}  
\\\midrule
	&
	&
	$7.98$ &
	$4.91$ &
	$52$ &
	$45$ &
	$16$ &
	$45$ &
	$3$ &
	$4$ &
	$1$ &
	$2$ &
	$4$ &
    $4$ &
    $42$ &
	$83$ &
	$4.0$  
\\ 
$\checkmark$ &
&
$5.06$ &
$3.58$ &
$98$ &
$77$ &
$28$ &
$79$ &
$51$ &
$55$ &
$29$ &
$37$ &
$17$ &
$8$ &
$\textbf{99}$ & 
$\textbf{100}$ & 
$2.20$  
\\ 
&
$\checkmark$ &
$1.02$ &
$1.49$ &
$95$ &
$87$ &
$58$ &
$90$ &
$70$ &
$76$ &
$27$ &
$39$ &
$48$ &
$52$ &
$98$ & 
$99$ & 
$0.67$ 
\\ 
$\checkmark$ &
$\checkmark$ &
$\textbf{0.93}$ &
$\textbf{1.40}$ &
$\textbf{99}$ &
$\textbf{92}$ &
$\textbf{60}$ &
$\textbf{91}$ &
$\textbf{74}$ &
$\textbf{77}$ &
$\textbf{43}$ &
$\textbf{53}$ &
$\textbf{51}$ &
$\textbf{56}$ &
$\textbf{99}$ & 
$\textbf{100}$ & 
$\textbf{0.62}$  
\\ 

\bottomrule

\end{tabular}
\vspace{0.2cm}
\caption{\textbf{Input Ablation:} Model performance by input. First row computed from dataset statistics.}
\label{table:inputs_ablation}
\vspace{-0.5cm}
\end{table} 
\paragraph{Continuous Learning:}
\label{experiments:continuous_learning}
In a real-world setting, the tag attributes supported by our system will grow as requirements change, and new supervised data becomes available. Ideally, models should support continuous learning, in which new tag attributes are trained on top of a learned model. 
Therefore, we explore whether our learned embedding can generalize to new attributes.
In this experiment, we add an additional set of 4 attributes to create a new set $\attrs_{new}$. We introduce two attributes to tag whether a vehicle \textit{causes} another vehicle to be blocked or braking and two attributes to tag whether a vehicle is specifically causing the SDV to be blocked or braking. 
We train our previously converged model with $\attrs_{\text{new}}$ for an additional 25 epochs and compare performance against an independently trained model for each new attribute. 
Fig.~\ref{fig:continuous_learning_tsne} shows that
our model always outperforms the independently trained baseline and 
can provide up to \textbf{45\%} increase in F1 score. 
Importantly, our continuously trained model maintained similar performance on all other attributes. 

\subsection{Ablation Studies and Visualization}
For the following experiments, we evaluate the performance of each tag attribute at its relevant locations (e.g., everywhere for actor density, locations with vehicles for actions, intersections for map topology and locations with moving vehicles for speed.) 

\paragraph{Joint vs. Separate Model:} 
In Fig.~\ref{fig:dimension_ablation_and_task_balancing}, we compare the performance of our model trained with different embedding dimensions and configurations. 
For each dimension, we compare our universal embedding versus an approach with separated embeddings for each tag category, each with an independently trained model. Despite requiring \textit{one fourth} the model capacity, our approach significantly outperforms the separated approach when embedding memory is limited ($D_e = 4, 16$) as it learns to share the embedding space more efficiently. Given a large enough  embedding ($D_e = 64$), both approaches perform similarly. The separated approach, however, does not scale as it requires both large embedding memory and a new network to be trained for each new tag category.

\paragraph{Leveraging Inputs:} 
In Table~\ref{table:inputs_ablation}, we ablate model inputs to understand the relative gain from adding LiDAR and HD maps.
The first row uses no input, outputting tag values based on dataset statistics. 
L1 error is minimized by predicting the tag's median and F1 score is maximized by always predicting positive.
Unsurprisingly, the poor performance of this baseline demonstrates that our model cannot simply exploit dataset biases and must utilize sensor input. 
Overall, we find maximal performance is attained when leveraging both LiDAR and HD maps. 
However, we find that HD map information alone is sufficient to tag map topology attributes and parked vehicles. 
Similarly, vehicle speed and turns achieve strong performance with only LiDAR as these attributes are primarily motion-based.

\paragraph{Task Balancing:} 
Fig.~\ref{fig:dimension_ablation_and_task_balancing} plots validation metrics throughout training with and without our task balancing scheme. 
We notice that our approach to sampling examples converges in fewer epochs and achieves better performance across both continuous and discrete tag attributes. 

\paragraph{Visualizing Embeddings:} Fig.~\ref{fig:continuous_learning_tsne} visualizes the embeddings of evaluation logs using t-SNE \cite{maaten2008visualizing}, considering spatial locations in the embedding tensor $\M E$ that are associated with vehicles. Three dominant clusters are apparent: parked vehicles (blue), stopped vehicles (green), and vehicles keeping lane (orange). Additional clusters are present for left and right turns. This demonstrates that the embedding captures high level semantics about each spatial location (e.g., actions at that location).

\vspace{-0.1in}

\section{Conclusion}
\label{sec:conclusion}
\vspace{-0.15in}
We introduced a novel dataset and approach for spatio-temporal tagging of self-driving scenes. Our model's output can be directly applied to generating interpretable scene tags and data log retrieval.
Our approach is designed to be efficient through the use of a universal embedding, and achieves good performance across a diverse set of tags.
Additionally, we demonstrate that our embedding is generalizable and can be used to improve the performance of new tags. 
We plan to release a benchmark, as we believe there are many exciting directions for future work including model improvements, more complex compositional tags and additional sensor inputs (e.g., RGB cameras, Radar).
 
\clearpage

\bibliography{paper}  %

\end{document}